\documentclass[runningheads]{llncs}
\usepackage[T1]{fontenc}
\usepackage{graphicx}
\usepackage{amsmath}
\usepackage{amssymb}
\usepackage{booktabs}
\usepackage{makecell}
\usepackage{multirow}
\usepackage{url}
\usepackage{hyperref}
\providecommand{\doi}[1]{\url{https://doi.org/#1}}
\usepackage{color}

\begin{document}
\title{THGFM: Dual-Branch Temporal Heterogeneous Graph Fusion Model}

\author{
Yixin Peng\inst{1}\thanks{Corresponding author.}
\and
Diego Collarana\inst{2}
\and
Er Jin\inst{1}
\and
Stefan Decker\inst{1,2}
}

\authorrunning{Y. Peng et al.}

\institute{
RWTH Aachen University, Aachen, Germany\\
\email{\{peng,jin,decker\}@dbis.rwth-aachen.de}
\and
Fraunhofer FIT, Sankt Augustin, Germany\\
\email{diego.collarana.vargas@fit.fraunhofer.de}
}
\maketitle              
\begin{abstract}
Temporal heterogeneous graphs offer a natural abstraction for dynamic relational systems in which diverse node and relation types co-exist and evolve over time.
Learning on such graphs requires jointly modeling cross-type
structural heterogeneity and the temporal dynamics of interactions, yet existing methods still struggle to reconcile parameter-efficient
cross-type transfer with relation-aware specialization, and typically inject time only as additive features outside the attention kernel.
We propose \textbf{THGFM}, a web-scale temporal heterogeneous graph
fusion model that addresses both limitations within a unified dual-path architecture. 
THGFM couples a \textit{Shared-Space Temporal Attention} branch for parameter-efficient cross-type transfer with a \textit{Relational Type-Partitioned Temporal Attention} branch for relation-aware specialization, and integrates them through \textit{Dual-Path Relational--Shared Fusion}, instantiated with \textit{Type-Conditioned Non-Competitive Gated Sum Fusion}: a adaptive mechanism that assigns independent, type-conditioned feature-wise gates to the shared and specialized branches, allowing both to be amplified or suppressed without zero-sum competition. 
To directly incorporate relative time into the attention score, THGFM further introduces \textit{Rotary Temporal Attention}, which rotates queries and keys by half-phases of relative time before matching. 
THGFM consistently outperforms baseline graph transformer models on academic graphs benchmarks, delivering a $+3.25\%$ six-task mean gain, with peak relative gains of $+12.37\%$ on OAG-CS PV, $+4.87\%$ on PF-$L_2$, and $+1.18\%$ on PF-$L_1$, and $+4.24\%$, $+3.73\%$, and $+4.61\%$ on OGBN-MAG, HTAG-ArXiv, and HTAG-DBLP, respectively.
\keywords{Temporal Heterogeneous Graphs \and Graph Transformers \and Temporal Attention \and Representation Learning \and Graph Foundation Models}
\end{abstract}


\section{Introduction}
\label{sec:introduction}
\begin{figure}[t]
  \centering
  \includegraphics[width=0.6\linewidth]{\detokenize{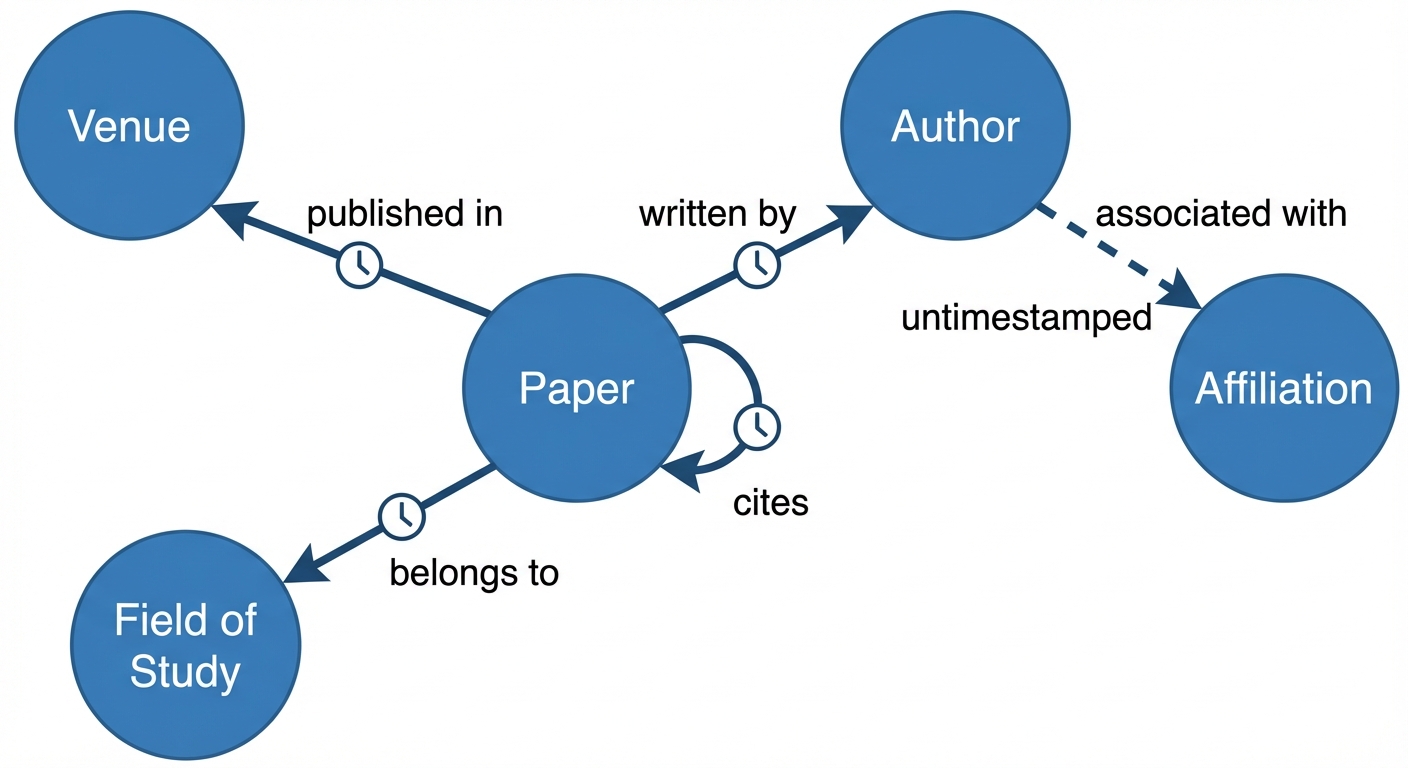}}
  \caption{OAG as a temporal heterogeneous graph with multiple node types and relation semantics that evolve over time. Timestamp availability is relation-dependent: publication-related edges are time-stamped by publication year, while \emph{author-affiliation} links in this example do not provide explicit timestamps and are therefore treated as edges with missing time (\(\psi(e)=\bot\)). The figure is adapted from \cite{Hu2020HGT}.}
  \label{fig:oag_temporal_example}
\end{figure}

A heterogeneous graph is a graph in which each node and edge is associated with a type, providing a natural representation for
real-world relational systems where diverse object types interact through multiple semantic relations. 
Representative examples include academic graphs, e-commerce graphs, scientific knowledge graphs, and enterprise information networks~\cite{Hogan2021KnowledgeGraphs,Niu2020DHGAT,Zhang2019OAG,Shang2023IndustrialKG}. 
As illustrated in Fig.~\ref{fig:oag_temporal_example}, an Open Academic Graph (OAG) instance comprises \emph{author}, \emph{paper}, \emph{venue}, and \emph{field-of-study} nodes linked by relations such as \emph{written-by}, \emph{published-in}, \emph{belongs-to}, \emph{associated-with} and \emph{cites}. 
These relation types encode different semantics, while node types exhibit markedly different feature distributions. 
Moreover, OAG is inherently temporal: new papers and collaborations appear continuously, and the context of existing entities changes as surrounding publications and links evolve over time. 
Modeling such temporal heterogeneous graphs, therefore, requires jointly capturing cross-type structural heterogeneity and temporal dynamics.

Recent heterogeneous graph \textbf{transformers} such as HGT~\cite{Hu2020HGT}, together with emerging graph foundation models~\cite{BechlerSpeicher2026GraphBFF}, have significantly advanced graph representation learning. 
Yet temporal heterogeneous graph learning remains much less developed than the broader dynamic graph learning researches
~\cite{Feng2024DynamicGNN}, with only a limited number of methods
designed specifically for temporal heterogeneous graphs, including
HTGNN~\cite{Fan2021HTGNN}, CTRL~\cite{Li2024CTRL}, and THAN~\cite{Li2023THAN}. 
Two challenges are particularly important for temporal heterogeneous graphs learning. 
First, heterogeneous modeling involves two design choices that are not competing alternatives but \emph{complementary} and jointly necessary: type- and relation-specific parameters, which preserve the fine-grained per-type and per-relation semantics that make heterogeneous graphs informative~\cite{Hong2020HetSANN,Hu2020HGT,Schlichtkrull2018RGCN,Wang2019HAN,Yu2022RHGNN}, and shared cross-type transformations, which enable parameter-efficient cross-type transfer and better generalization to rare node and relation types by amortizing statistics across the graph~\cite{He2024Heta,Lv2021SimpleHGN,Vashishth2020CompGCN}.
Recent work on billion-scale graph foundation models provides evidence that the
shared and specialized views are complementary in principle: within a single
heterogeneous attention block, combining type-agnostic and type-conditioned
attention can express graph functions that neither component alone can
realize~\cite{BechlerSpeicher2026GraphBFF}. 
This argues against dropping either side of the design and motivates architectures that keep both signals and learn how to combine them, rather than treating them as an expressivity-versus-efficiency trade-off. 
Second, temporal information is still commonly handled outside the attention kernel itself. HGT and later temporal heterogeneous graph models such as HTGNN, THAN, and CTRL already incorporate temporal signals, but they do so mainly through additive encodings, across-time aggregation, or continuous-time influence modeling.
More specifically, in HGT, Relative Temporal Encoding (RTE) influences attention only indirectly
through the source representation. For temporal heterogeneous graphs,
however, relative time should affect not only what message is sent, but which neighbors deserve attention.

On OAG, for example, two papers appearing at a conference may discuss closely related topics, yet their predictive relevance to a target paper in citation or collaboration contexts usually decays as their publication years become more distant. 
Therefore, temporal separation should reshape which neighbors receive high attention, not only the encoding of the messages they send.

Moreover, recent graph models have shown that adaptive fusion is useful when different information sources should contribute unequally across nodes. 
In heterogeneous graphs, HGAMLP~\cite{Liang2024HGAMLP} introduces a node-adaptive weight adjustment mechanism to fuse different knowledge sources, while AFMF~\cite{Liu2025AFMF} shows that globally shared fusion weights can become too rigid when neighborhood evidence varies across nodes. 
This issue is especially central to our setting: a dual-path temporal heterogeneous model is only effective if it can decide, for each target node, how much to rely on shared cross-type transfer versus relation-aware specialization. 
In temporal heterogeneous graphs, both are necessary, but their relative importance varies across node types, neighborhood structures, and temporal contexts. 
We therefore foreground dual-branch fusion as a principal
architectural decision rather than a post hoc aggregation step, and
introduce \emph{Type-Conditioned Non-Competitive Gated Sum Fusion}
(TC-NGSF), which learns two independent feature-wise gates conditioned jointly on the paired branch representations and the target-node type, and then combines the recalibrated branches through summation.
Unlike competitive fusion, TC-NGSF does not force the two branches into a zero-sum trade-off, allowing THGFM to amplify or suppress shared and specialized evidence independently when both are informative.

These observations motivate THGFM, a \emph{Temporal Heterogeneous Graph
Fusion Model} (Figure~\ref{fig:hgfm_overall}), a transformer-based architecture that combines a \emph{Shared-Space Temporal Attention}
(SSTA) branch, which improves parameter sharing and cross-type
transfer, with a \emph{Relational Type-Partitioned Temporal Attention} (RTTA)
branch, which preserves type- and relation-aware semantics. The two
branches are integrated through \emph{Dual-Path Relational--Shared
Fusion} (DRSF), where TC-NGSF serves as the core adaptive fusion mechanism of the
full THGFM model. To model graph
dynamics more directly, THGFM further introduces \emph{Rotary Temporal
Attention} (RoTA), which injects edge-wise relative time into
attention by rotating the query and key with opposite half-phases
determined by the relative time interval before score computation,
which is inspired by the rotary-positioning principle of RoFormer~\cite{su2021roformer}. Hence THGFM is 
particularly well suited to temporal
heterogeneous graphs, where heterogeneity, dynamics, and distribution
imbalance must be handled simultaneously.

\paragraph{Contributions.}
\begin{enumerate}
    \item We propose \textbf{THGFM}, a dual-path temporal heterogeneous
    graph fusion model built on two complementary attention branches:
    \emph{SSTA} shares \(Q/K/V\) projections after type-specific
    adapters to support parameter-efficient cross-type transfer, while
    \emph{RTTA} preserves fine-grained semantics through target-type
    queries, source-type keys/values, and relation-grouped
    per-relation transforms.

    \item We introduce two operators tailored to temporal heterogeneous
    fusion. \emph{RoTA} injects edge-wise relative time directly into
    attention scoring by rotating queries and keys with opposite
    half-phases of the time interval; \emph{TC-NGSF} combines the two
    branches through independent type-conditioned feature-wise gates and
    additive aggregation, avoiding zero-sum competition between them.

    \item Although RTTA is inspired by HGT, it reduces the per-layer edge
    cost from \(\mathcal{O}(|E||\mathcal{T}|^{2}|\mathcal{R}|)\) to
    \(\mathcal{O}(|E|(|\mathcal{T}|{+}|\mathcal{R}|))\), while retaining
    comparable accuracy to HGT and even improving performance on some
    tasks.

    \item On OAG-CS, OGBN-MAG, HTAG-ArXiv, and HTAG-DBLP, THGFM
    consistently outperforms HGT, attaining a \(+3.25\%\) six-task mean
    gain and up to \(+12.37\%\) peak; an ablation over branches,
    temporal switches, and six DRSF operators validates each design
    choice.
\end{enumerate}

The rest of the paper is organized as follows. Section~\ref{sec:related} reviews the most relevant literature. Section~\ref{sec:preliminaries} introduces the temporal heterogeneous graph setting and relative temporal encoding. Section~\ref{sec:method} presents the THGFM architecture and the proposed operators. Section~\ref{sec:experiments} describes the experimental setup and reports the empirical results. Section~\ref{sec:conclusion} concludes the paper.

\section{Related Work}
\label{sec:related}

We organize related work along three lines most relevant to THGFM:
heterogeneous graph representation learning, temporal graph learning and
temporal encoding, and adaptive fusion of complementary information
sources.

\subsection{Heterogeneous Graph Representation Learning}
\label{sec:heterogeneous_graph_representation_learning}

Work on heterogeneous graph learning has gradually moved from schema-guided designs to learned relation-aware models. 
Early methods
such as HAN~\cite{Wang2019HAN} and MAGNN~\cite{Fu2020MAGNN} rely on meta-paths to capture heterogeneous semantics. 
These models are effective when good semantic paths are known in advance, but their behavior still depends heavily on handcrafted structure. 
Later methods such as RGCN~\cite{Schlichtkrull2018RGCN} and
Simple-HGN~\cite{Lv2021SimpleHGN} reduce this dependence by learning relation-aware
transformations or heterogeneous attention directly from
data. More recent scalable baseline, including SeHGNN~\cite{Yang2023SeHGNN}, further shows that efficiency and
large-graph training are now central concerns in practical heterogeneous learning.

This progression highlights the primary motivation for THGFM: an effective model for heterogeneous graphs should preserve relation-specific semantics without relying on handcrafted meta-paths, while avoiding fully type-isolated modeling, which does not scale efficiently to large graphs. Thus, heterogeneous learning requires both specialization and controlled sharing. THGFM is designed to achieve this balance.

\subsection{Temporal Graph Learning and Temporal Encoding}
\label{sec:temporal_graph_learning}

Temporal graph learning asks how evolving interactions should change
representation learning beyond static structure alone. Representative
approaches include continuous-time attention models such as TGAT~\cite{Xu2020TGAT} and
memory-based event models such as TGN~\cite{Rossi2020TGN}.
These methods establish that temporal information is often essential,
but they are mainly designed for homogeneous graphs or settings with
limited type complexity.

For heterogeneous temporal graphs, HGT~\cite{Hu2020HGT} is the most relevant starting
point because it combines heterogeneous attention with Relative Temporal
Encoding (RTE). Later work such as HTGNN~\cite{Fan2021HTGNN} and CTRL~\cite{Li2024CTRL}
extend temporal modeling through across-time aggregation or
event-driven continuous-time mechanisms.
However, the dominant pattern is still to inject time through source
features, temporal histories, or auxiliary dynamics modules. In other
words, time affects attention indirectly. For temporal
heterogeneous graphs, this is not enough: temporal distance should
influence not only what information a neighbor carries, but also how
strongly that neighbor is matched by the attention kernel itself. This
is the gap targeted by our temporal design.

\subsection{Fusion of Complementary Information Sources}
\label{sec:fusion_between_two_modalities}

Adaptive fusion has become increasingly important when multiple
information sources provide complementary but uneven evidence across
nodes. In heterogeneous graph learning, HGAMLP~\cite{Liang2024HGAMLP}
uses node-adaptive weighting to combine different knowledge sources,
while AFMF~\cite{Liu2025AFMF} shows that globally shared fusion weights
can be overly rigid under diverse neighborhood patterns. These studies
establish the value of data-dependent fusion, but they are not designed
for the dual-path temporal heterogeneous setting considered here.

In our setting, fusion must reconcile two distinct inductive biases:
shared-space cross-type transfer and relation-aware specialization.
This requirement is more structured than standard feature-view fusion,
because the two branches should be allowed to be simultaneously strong
or simultaneously weak depending on node type, neighborhood structure,
and temporal context. THGFM addresses this regime with branch-level
type-conditioned non-competitive gating, which avoids forcing the two
branches into a zero-sum trade-off.




\section{Preliminaries}
\label{sec:preliminaries}

\subsection{Temporal Heterogeneous Graph Setting}
\label{sec:thg_setting}

We extend the definition of temporal heterogeneous graphs from HGT~\cite{Hu2020HGT} to include edge timestamps. 
A temporal heterogeneous graph is a directed typed graph
\(G^\star=(V^\star,E^\star,\tau,\phi,\psi)\), where
\(\tau:V^\star\to\mathcal{A}\) maps nodes to node types,
\(\phi:E^\star\to\mathcal{R}\) maps edges to relation types, and
\(\psi:E^\star\to\mathbb{Z}\cup\{\bot\}\) provides an optional timestamp per edge (e.g. Author-Affiliation
links do not provide explicit timestamps). For an edge \(e=(u,v)\) with relation
\(r=\phi(e)\), its meta relation is \(\langle \tau(u),r,\tau(v)\rangle\);
if reverse edges are explicitly included, we denote the inverse relation
by \(r^{-1}\). Edge timestamps are used whenever they are semantically well defined (e.g., publication-year on Paper-Field or Paper-Venue edges).
Relations without a timestamp are assigned \(\bot\). For training, each iteration operates on a locally sampled subgraph
\(G=(V,E)\) using HGSampling as in HGT~\cite{Hu2020HGT}. Each sampled
node \(v\in V\), when treated as a target node, is assigned an induced
context time \(t_v\in\mathbb{Z}\). When expanding along an incoming edge
\(e=(j,i)\) from target \(i\), the source-node time is assigned as
\begin{equation}
\label{eq:induced-context-time}
t_j \leftarrow
\begin{cases}
\psi(e), & \psi(e)\neq\bot,\\
t_i, & \psi(e)=\bot.
\end{cases}
\end{equation}
Following the HGSampling procedure of HGT, the assignment in
Eq.~\eqref{eq:induced-context-time} preserves temporal
consistency along each sampled path: timestamped edges propagate their
own event time, while non-timestamped edges inherit the current target
context time. A side effect is that the same global node may receive
different induced times across sampled batches. We keep this assignment
rule aligned with HGT to ensure a fair baseline comparison under matched
sampling semantics, so observed improvements primarily reflect the model
architecture rather than a changed temporal propagation protocol.

After tensorization, we relabel sampled nodes to local indices
\(\{0,\ldots,|V|-1\}\). Let \(N=|V|\) and \(M=|E|\). We construct
\(\mathbf{X}\in\mathbb{R}^{N\times d_{\mathrm{in}}}\) (node features),
\(\boldsymbol{\tau}\in\mathcal{A}^{N}\) (node-type IDs), and an edge
index \(\mathbf{A}\in\{0,\ldots,N-1\}^{2\times M}\), whose \(e\)-th
column stores a directed edge \((j_e,i_e)\). In the same edge order, we
store relation IDs \(\boldsymbol{\phi}\in\mathcal{R}^{M}\) and define
edge-wise relative time from induced node times:
\begin{equation}
\label{eq:delta_t_from_node_times}
\Delta t_{ji} = t_i - t_j + c,
\end{equation}
where \(c\in\mathbb{Z}\) is a fixed offset chosen so that values are
non-negative. Equivalently, for edge slot \(e\),
\(\Delta t_e = t_{i_e}-t_{j_e}+c\), yielding
\(\boldsymbol{\Delta t}\in\mathbb{Z}_{\ge 0}^{M}\). Hence
\(\boldsymbol{\Delta t}\) is an edge-aligned input used in message
passing, computed on-the-fly from batch-local induced node times rather
than stored as a static global field. The tuple
\((\mathbf{X}, \boldsymbol{\tau}, \mathbf{A}, \boldsymbol{\phi}, \boldsymbol{\Delta t})\)
is the input of THGFM.

\subsection{Relative Temporal Encoding}
\label{sec:rte}
Following HGT, we use the relative interval defined in
Eq.~\eqref{eq:delta_t_from_node_times}. Given source representation
\(h_j\), Relative Temporal Encoding (RTE)
injects time through additive sinusoidal features:
\begin{equation}
\label{eq:rte}
\tilde h_{j\to i}
= h_j + \mathbf{W}_t\operatorname{PE}\!\left(\Delta t_{ji}\right),
\end{equation}
where \(\operatorname{PE}(\cdot)\) is sinusoidal positional encoding and
\(\mathbf{W}_t\) is a learnable projection. This source-side temporal
augmentation is used during edge-wise message construction.

\subsection{Rotary Temporal Attention}
\label{sec:rota}
Besides additive temporal encoding, THGFM further uses Rotary Temporal
Attention (RoTA) so that relative time also affects the matching stage
of attention. For each head \(m\), let \(q_i^{(m)}\) and \(k_{j\to i}^{(m)}\) denote the query
and key on edge \((j,r,i)\). We apply an edge-dependent rotary transform
\(\operatorname{Rot}(\cdot,\theta_{ji})\), where \(\theta_{ji}\propto \Delta t_{ji}\),
and compute attention from the rotated vectors. The symmetric rotation is
\begin{equation}
\label{eq:rota}
\begin{aligned}
\bar q_i^{(m)}
&=\operatorname{Rot}\!\left(q_i^{(m)},+\theta_{ji}/2\right),\\
\bar k_{j\to i}^{(m)}
&=\operatorname{Rot}\!\left(k_{j\to i}^{(m)},-\theta_{ji}/2\right).
\end{aligned}
\end{equation}
Attention is then scored by the dot product
\(\langle \bar q_i^{(m)}, \bar k_{j\to i}^{(m)}\rangle\). In this way,
temporal intervals act directly on attention score computation.

\section{The Temporal Heterogeneous Graph Fusion Model}
\label{sec:method}

\begin{figure}[t]
  \centering
  \includegraphics[width=\linewidth]{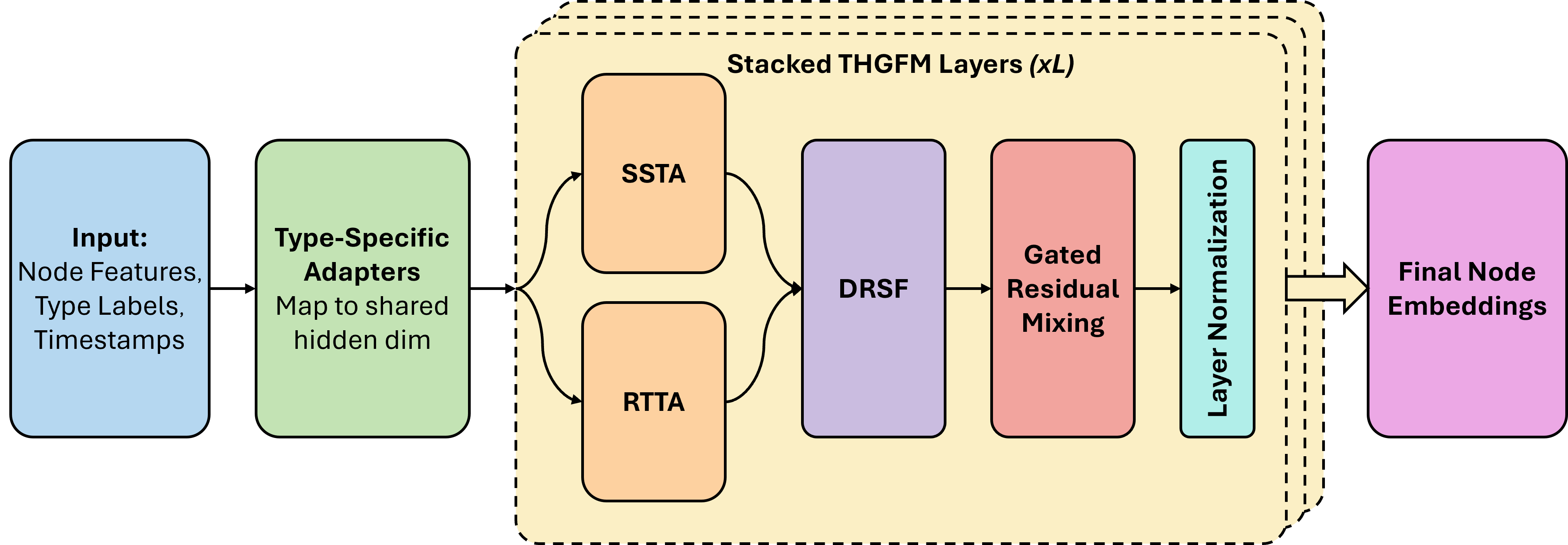}
  \caption{Overall architecture of THGFM. Each layer executes two temporal heterogeneous attention branches---SSTA (Shared-Space Temporal Attention) and RTTA (Relational Type- Partitioned Temporal Attention)---both modulated by Rotary Temporal Attention (RoTA), and fuses their outputs through Dual-Path Relational--Shared Fusion (DRSF) before passing to the next layer via a gated residual connection.}
  \label{fig:hgfm_overall}
\end{figure}

Building on the preliminaries, THGFM is a dual-branch temporal
heterogeneous graph fusion model. It combines an SSTA branch for
shared-space cross-type transfer and an RTTA branch for
relation-aware specialization, integrates them via DRSF, and injects
edge-wise temporal bias through RoTA.
Figure~\ref{fig:hgfm_overall} gives an overview of this architecture.

\subsection{Overall THGFM Architecture}
\label{sec:overall_hgfm}

As defined in Section~\ref{sec:thg_setting}, THGFM receives the node
features, node types, edge indices, relation IDs, and relative times of a
sampled temporal heterogeneous subgraph. The model first maps the raw
heterogeneous features to a shared hidden space via type-specific input
adapters. For node \(v\), the initial representation is
\begin{equation}
\label{eq:input-adapter}
\mathbf{h}^{(0)}_v
= \operatorname{Dropout}\!\left(
\tanh\!\left(\mathbf{W}^{\mathrm{in}}_{\tau(v)}\mathbf{x}_v\right)
\right),
\end{equation}
where \(\mathbf{W}^{\mathrm{in}}_{\tau(v)} \in \mathbb{R}^{d \times d_{\mathrm{in}}}\)
is a learnable type-specific input projection.

Representations are then updated by \(L\) stacked THGFM layers:
\begin{equation}
\label{eq:thgfm-layer-stack}
\mathbf{h}^{(\ell)}_v
=
\operatorname{THGFM}^{(\ell)}
\!\left(
\{\mathbf{h}^{(\ell-1)}_u\}_{u \in V},
\boldsymbol{\tau},
\mathbf{A},
\boldsymbol{\phi},
\boldsymbol{\Delta t}
\right),
\qquad \ell = 1, \dots, L.
\end{equation}
At each layer, attention-based heterogeneous message passing uses hidden
states together with node types, relation types, graph structure, and
relative time. Thus, one layer aggregates one-hop neighborhoods, and
the stack in Eq.~\eqref{eq:thgfm-layer-stack} yields an effective
\(L\)-hop receptive field on the
sampled subgraph. When DRSF is enabled, SSTA and RTTA are both computed
and fused \emph{within} the same layer before the residual update, i.e.,
THGFM adopts layer-wise fusion rather than late fusion over the final
branch outputs.

Each THGFM layer has two stages:
(i) \textit{message construction and attention normalization} over
incoming edges (including intra-layer SSTA--RTTA fusion when DRSF is
used), which produces a pre-residual representation
\(\mathbf{z}_i^{(\ell)}\in\mathbb{R}^{d}\) for each target node \(i\); and
(ii) \textit{gated residual mixing with type-specific normalization}.
Let \(\mathbf{h}_i^{(\ell-1)}\) denote the layer input for node \(i\).
The update rule is
\begin{equation}
\label{eq:type-gated-residual}
\begin{aligned}
\alpha_{\tau(i)}
&=\sigma\!\left(g_{\tau(i)}\right),\\
\mathbf{r}_i^{(\ell)}
&=
\alpha_{\tau(i)}\mathbf{z}_i^{(\ell)}
+
\bigl(1-\alpha_{\tau(i)}\bigr)\mathbf{h}_i^{(\ell-1)},\\
\mathbf{h}_i^{(\ell)}
&=\operatorname{LayerNorm}_{\tau(i)}\!\left(\mathbf{r}_i^{(\ell)}\right).
\end{aligned}
\end{equation}
The update in Eq.~\eqref{eq:type-gated-residual} is shared across
single-branch (SSTA-only or RTTA-only) and
DRSF settings: only \(\mathbf{z}_i^{(\ell)}\) changes with the message
block, while residual and normalization remain identical. Here
\(g_{\tau(i)}\) is a learnable scalar gate per target type,
\(\sigma\) is the logistic function, and
\(\alpha_{\tau(i)}\in(0,1)\) controls interpolation between
\(\mathbf{z}_i^{(\ell)}\) and \(\mathbf{h}_i^{(\ell-1)}\).

\subsection{Dual-Path Relational--Shared Fusion (DRSF)}
\label{sec:drsf}

\begin{figure*}[t]
  \centering
  \includegraphics[width=\linewidth]{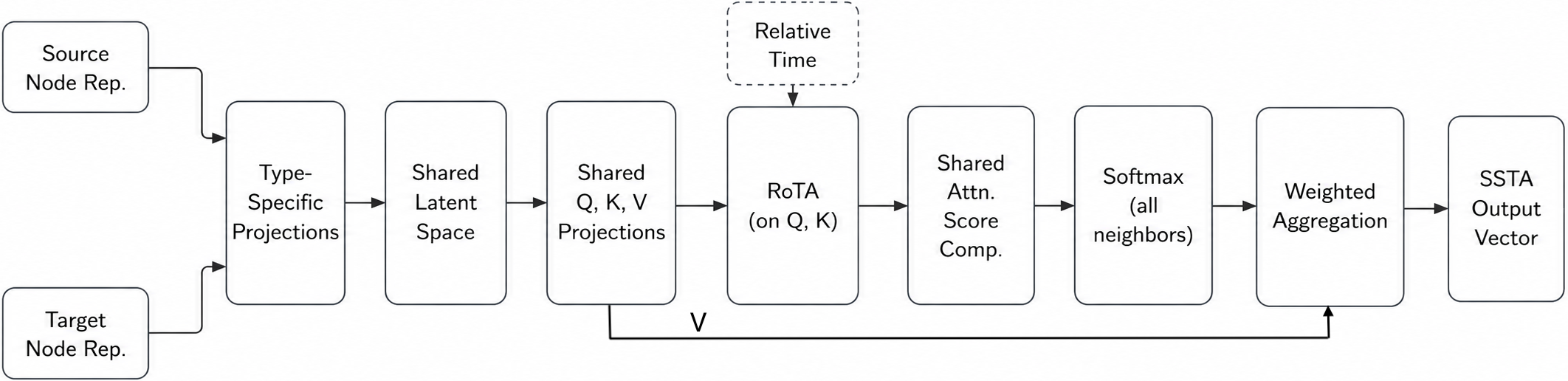}
\par\smallskip
{\small\textbf{(a) SSTA}. Type-specific input projections first map source/target features into a shared latent space, followed by shared \(Q/K/V\) projections. RoTA rotates \(q,k\) with edge-wise relative time before score computation, and attention is globally normalized across all incoming neighbors.}

  \vspace{0.8em}

  \includegraphics[width=\linewidth]{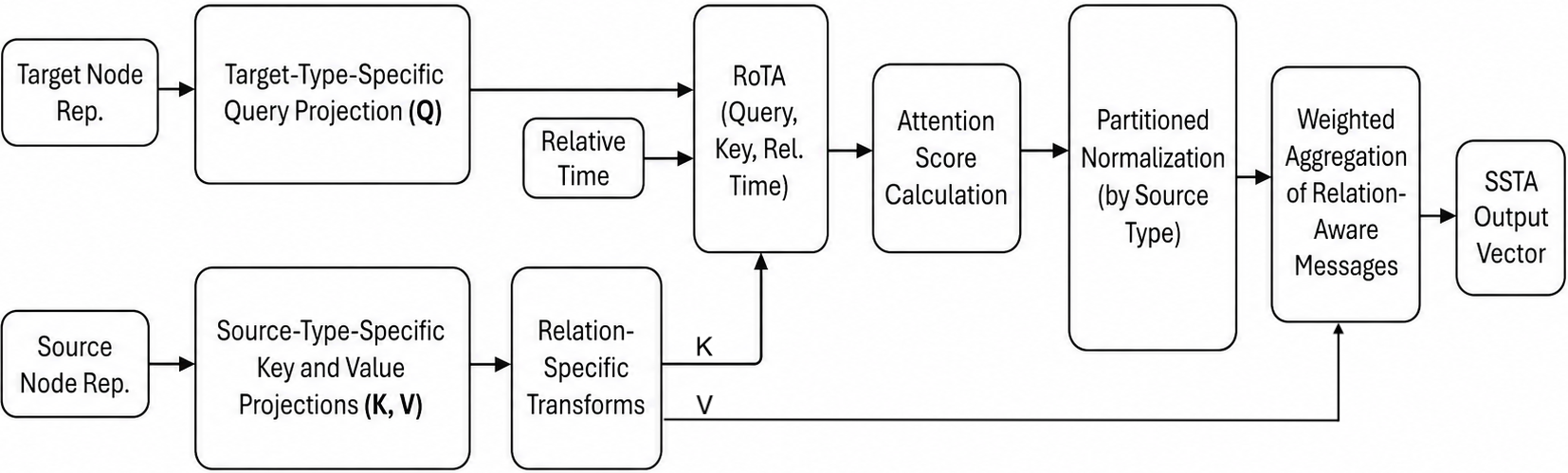}
\par\smallskip
{\small\textbf{(b) RTTA}. Target-type-specific queries and source-type-specific keys/values are further modulated by relation-specific transforms (\(A_r,M_r\)). After RoTA, attention is normalized within each source-type partition.}
  \caption{Two THGFM attention branches used by DRSF. SSTA provides parameter-efficient cross-type transfer in a shared space, while RTTA preserves type- and relation-aware specialization; both apply RoTA before attention scoring.}
  \label{fig:ssta_rtta}
\end{figure*}

DRSF combines two branches with complementary inductive roles:
\emph{SSTA} for shared-space cross-type transfer and \emph{RTTA} for
relation-aware heterogeneous interactions. The two branches are fused
within each THGFM layer before the target-side residual update, and the
fused representation serves as the layer output.
Figures~\ref{fig:ssta_rtta} and~\ref{fig:drsf_fusion} illustrate the two
branches and the fusion module, respectively.

\subsubsection{Shared-Space Branch: SSTA}
\label{sec:ssta}
SSTA first maps source and target representations into a shared latent
space via type-specific projections:
\begin{equation}
\label{eq:ssta-type-projections}
z_{j\to i}=W^{\text{type}}_{\tau(j)}\tilde h_{j\to i},
\qquad
z_i=W^{\text{type}}_{\tau(i)}h_i.
\end{equation}
It then applies shared \(Q/K/V\) projections to obtain the per-head
query, key, and value vectors \(q_i^{(m)}\), \(k_{j\to i}^{(m)}\), and
\(v_{j\to i}^{(m)}\), respectively.
After RoTA phase-rotates \(q_i^{(m)}\) and \(k_{j\to i}^{(m)}\) with
edge-wise relative time, attention is normalized \emph{globally} over
all neighbors regardless of type:
\begin{equation}
\label{eq:ssta-attention}
\alpha_{ji}^{(m)}
=\operatorname{softmax}_{j\in\mathcal N(i)}
\!\left(\tfrac{\langle \bar q_i^{(m)},\bar k_{j\to i}^{(m)}\rangle}{\sqrt{d_k}}\right),
\end{equation}
so neighbors of different source types compete in a single attention domain.
\subsubsection{RTTA Branch}
\label{sec:rtta}
RTTA instead uses target-type-specific queries, source-type-specific
keys/values, and relation-specific transforms:
\begin{equation}
\label{eq:rtta-projections}
\begin{aligned}
q_i^{(m)}
&=W_{\tau(i)}^{Q,(m)}h_i,\\
\hat k_{j\to i}^{(m)}
&=A_r^{(m)}W_{\tau(j)}^{K,(m)}\tilde h_{j\to i},\\
\hat v_{j\to i}^{(m)}
&=M_r^{(m)}W_{\tau(j)}^{V,(m)}\tilde h_{j\to i}.
\end{aligned}
\end{equation}
RoTA is applied to \(q_i^{(m)}\) and \(\hat k_{j\to i}^{(m)}\) as in SSTA.
Unlike the global normalization in Eq.~\eqref{eq:ssta-attention}, RTTA
normalizes attention \emph{within each source-type partition}:
\begin{equation}
\label{eq:rtta-attention}
\alpha_{ji}^{(m)}
=\operatorname{softmax}_{k\in\mathcal N_{\tau(j)}(i)}
\!\left(\tfrac{\langle \bar q_i^{(m)},\bar k_{j\to i}^{(m)}\rangle}{\sqrt{d_k}}\right).
\end{equation}

\begin{figure}[t]
  \centering
  \includegraphics[width=\linewidth]{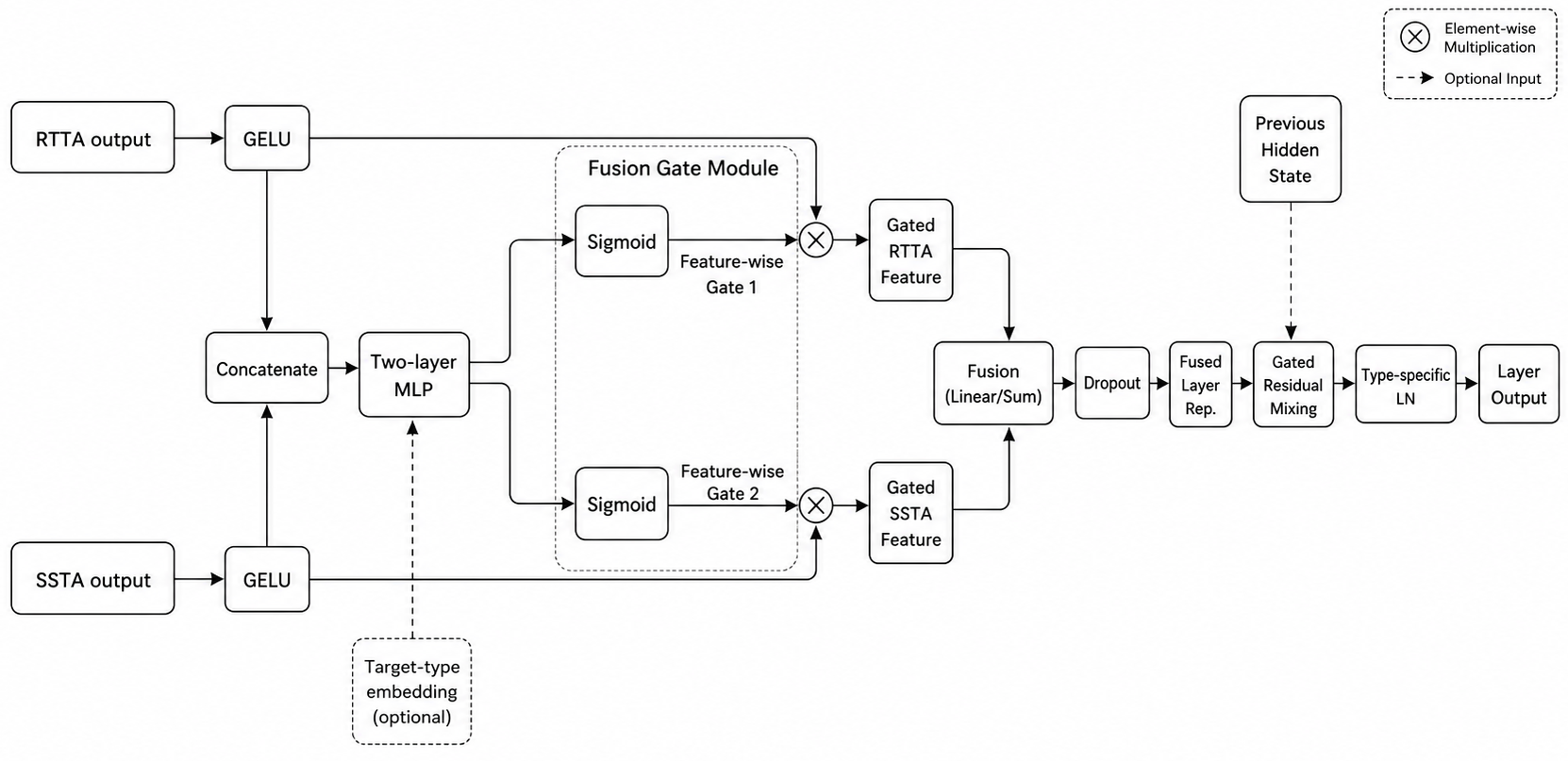}
  \caption{\textbf{DRSF}. Branch outputs $z_i^{\text{ssta}}$ and $z_i^{\text{rtta}}$ are activated with GELU and fed into one of six fusion operators. Gated fusion operators (NGLF, NGSF, TC-NGLF, TC-NGSF) compute non-competitive sigmoid gates from concatenated branch features---optionally conditioned on a target-type embedding $e_{\tau(i)}$---to recalibrate each branch before linear projection or summation.}
  \label{fig:drsf_fusion}
\end{figure}

\subsubsection{Fusion in DRSF}

Within a THGFM layer, DRSF executes the SSTA and RTTA branches in parallel
and fuses their outputs before the type-wise residual update (Figure~\ref{fig:drsf_fusion}).
RoTA is applied in both branches before attention-score computation. Let
\(z_i^{\text{rtta}},z_i^{\text{ssta}}\in\mathbb{R}^d\) denote the
pre-fusion aggregated outputs for target node \(i\), matching the split
of the message-passing aggregate before fusion. We apply GELU to obtain
\(u_i^{\text{rtta}}=\operatorname{GELU}(z_i^{\text{rtta}})\) and
\(u_i^{\text{ssta}}=\operatorname{GELU}(z_i^{\text{ssta}})\), and then
compare six DRSF operators: SF, LF, NGLF, NGSF, TC-NGLF, and TC-NGSF.

\paragraph{Ungated fusion (LF and SF).}
The two ungated operators are
\begin{equation}
\label{eq:ungated-fusions}
\begin{aligned}
f_i^{\text{LF}}
&=\operatorname{Dropout}\!\left(
W^{\text{fuse}}[u_i^{\text{rtta}};u_i^{\text{ssta}}]
\right),\\
f_i^{\text{SF}}
&=\operatorname{Dropout}\!\left(
u_i^{\text{rtta}}+u_i^{\text{ssta}}
\right).
\end{aligned}
\end{equation}

\paragraph{Non-competitive gating.}
\label{sec:drsf-gated-shared}
Four gated fusion operators (NGLF, NGSF, TC-NGLF, TC-NGSF) compute
two feature-wise sigmoid gates without cross-branch competition.
Their gate inputs, gates, and recalibrated features are jointly defined as
\begin{equation}
\label{eq:noncompetitive-gating}
\begin{aligned}
\gamma_i
&=
\begin{cases}
[u_i^{\text{rtta}};u_i^{\text{ssta}}],
& \text{NGLF/NGSF},\\
[u_i^{\text{rtta}};u_i^{\text{ssta}};e_{\tau(i)}],
& \text{TC-NGLF/TC-NGSF},
\end{cases}\\
[g_i^{\text{rtta}};g_i^{\text{ssta}}]
&=
\operatorname{sigmoid}\!\left(
W_2\,\operatorname{GELU}\!\left(W_1\gamma_i+b_1\right)+b_2
\right),\\
\tilde{t}_i^{\text{rtta}}
&=g_i^{\text{rtta}}\odot u_i^{\text{rtta}},
\qquad
\tilde{t}_i^{\text{ssta}}
=g_i^{\text{ssta}}\odot u_i^{\text{ssta}}.
\end{aligned}
\end{equation}
Here, \(\gamma_i\in\mathbb{R}^{2d}\) for NGLF/NGSF and
\(\gamma_i\in\mathbb{R}^{2d+d_\tau}\) for their type-conditioned variants;
\(e_{\tau(i)}\in\mathbb{R}^{d_\tau}\) is a learnable target-type embedding,
and \(g_i^{\text{rtta}},g_i^{\text{ssta}}\in\mathbb{R}^{d}\).
The parameters \(W_1,W_2,b_1,b_2\) use the corresponding input dimension.

\paragraph{Gated fusion (NGLF and NGSF).}
Using the recalibrated features in Eq.~\eqref{eq:noncompetitive-gating},
NGLF applies a linear projection, whereas NGSF uses summation:
\begin{equation}
\label{eq:gated-fusions}
\begin{aligned}
f_i^{\text{NGLF}}
&=
\operatorname{Dropout}\!\left(
W^{\text{fuse}}
\left[
\tilde{t}_i^{\text{rtta}};
\tilde{t}_i^{\text{ssta}}
\right]
\right),\\
f_i^{\text{NGSF}}
&=
\operatorname{Dropout}\!\left(
\tilde{t}_i^{\text{rtta}}+\tilde{t}_i^{\text{ssta}}
\right).
\end{aligned}
\end{equation}

\paragraph{Type-conditioned variants (TC-NGLF and TC-NGSF).}
These variants use the second gate input in
Eq.~\eqref{eq:noncompetitive-gating}. TC-NGLF and TC-NGSF otherwise follow
the first and second lines of Eq.~\eqref{eq:gated-fusions}, respectively;
their outputs are denoted by \(f_i^{\text{TC-NGLF}}\) and
\(f_i^{\text{TC-NGSF}}\).

All six DRSF fusion operators (SF, LF, NGLF, NGSF, TC-NGLF, and TC-NGSF)
share the same type-wise residual update described
in Section~\ref{sec:overall_hgfm}.

\section{Evaluation}
\label{sec:experiments}

\subsection{Web-Scale Datasets.}
We conduct experiments on temporal heterogeneous academic benchmarks built from both large-scale academic graph sources and curated HTAG~\cite{Liu2024HTAG} releases.
OAG-CS is derived from the Computer Science subgraph of the Open Academic Graph and Microsoft Academic Graph raw datasets~\cite{Wang2020MAG,Zhang2019OAG}.
HTAG-ArXiv and HTAG-DBLP are drawn from the HTAG datasets~\cite{Liu2024HTAG}.
These benchmarks are modeled as temporal heterogeneous graphs with typed nodes, typed directed edges, and edge times.
Table~\ref{tab:dataset_statistics_extended} summarizes the scale and
relation composition of these benchmarks.


\begin{table*}[t]
\caption{Statistics of benchmarks.}
\centering
\footnotesize
\setlength{\tabcolsep}{6pt}
\renewcommand{\arraystretch}{1.12}
\noindent\textbf{(a) Entity-scale statistics.}\par\smallskip
{\setlength{\tabcolsep}{4.5pt}%
\resizebox{\linewidth}{!}{%
\begin{tabular}{@{}l r r r r r r r@{}}
\toprule
\textbf{Dataset}
& \textbf{\#Nodes} & \textbf{\#Edges} & \textbf{\#Papers} & \textbf{\#Authors} & \textbf{\#Fields} & \textbf{\#Venues} & \textbf{\#Affiliation} \\
\midrule
OAG-CS
& 11,732,027 & 107,263,811 & 5,597,605 & 5,985,759 & 119,537 & 27,433 & 16,931 \\
OGBN-MAG
& 1,940,092 & 27,196,251 & 736,389 & 1,134,649 & 59,965 & 349 & 8,740 \\
HTAG-ArXiv
& 231,151 & 2,075,692 & 81,634 & 127,590 & 21,887 & -- & -- \\
HTAG-DBLP
& 1,989,019 & 29,830,033 & 964,350 & 958,961 & 65,699 & -- & -- \\
\bottomrule
\end{tabular}}}
\par\medskip
\noindent\textbf{(b) Relation counts by Meta-path (typed edges).}\par\smallskip
\resizebox{\linewidth}{!}{%
\begin{tabular}{@{}l r r r r r@{}}
\toprule
\textbf{Dataset}
& \textbf{\#P--A} & \textbf{\#P--F} & \textbf{\#P--V} & \textbf{\#A--I} & \textbf{\#P--P} \\
\midrule
OAG-CS
& 15,571,614 & 47,462,559 & 5,597,606 & 7,190,480 & 31,441,552 \\
OGBN-MAG
& 7,145,660 & 7,505,078 & 736,389 & 1,043,998 & 10,765,126 \\
HTAG-ArXiv
& 300,233 & 755,835 & -- & -- & 1,019,624 \\
HTAG-DBLP
& 3,070,343 & 10,080,164 & -- & -- & 16,679,526 \\
\bottomrule
\end{tabular}}
\label{tab:dataset_statistics_extended}
\end{table*}
\subsection{Experimental Setup}
\label{sec:exp_setup}

\paragraph{Tasks.}
We evaluate THGFM on four temporal academic benchmarks. On OAG-CS, we
consider paper--field prediction at two granularities (\(L_1\), \(L_2\))
and paper--venue prediction~\cite{Zhang2019OAG,Hu2020HGT}. On OGBN-MAG~\cite{hu2020ogb}, we
perform paper--venue classification over 349 venues. HTAG-ArXiv and
HTAG-DBLP~\cite{Liu2024HTAG} are multi-class node-classification tasks with
dataset-specific label semantics (40 arXiv CS subject areas and 9 DBLP
topic labels, respectively). Temporal splits follow those benchmark protocols.


\paragraph{Dataset Preprocessing.}
We use our temporal heterogeneous preprocessing pipeline across OAG-CS,
OGBN-MAG, HTAG-DBLP, and HTAG-ArXiv, with a consistent schema and
split-aware supervision. We build typed graphs over paper, author, field,
and plus venues and affiliation when available. Relations include
paper--paper, paper--field, paper--venue, and author--paper edges. Paper
features follow benchmark-native sources (OAG
XLNet~\cite{yang2019xlnet}, OGB word2vec~\cite{church2017word2vec}, HTAG released
dense embeddings), while non-paper features are propagated from paper
neighborhoods.

All datasets are converted to a shared intermediate representation from
which subprocesses materialize only batch-local subgraphs at training time.
Compared with pyHGT preprocessing~\cite{Hu2020HGT}, this pipeline enforces stricter
temporal consistency and numerically stable structural features,
improving memory efficiency and reducing dataset-specific engineering
variance.

\begin{table}[t]
\caption{\textbf{(a)} HGT vs.\ our preprocessing (OAG-CS only). \textbf{(b)} Peak RAM, offline graph prep. time, and per-epoch wall time for the same HGT setting, training recipe and with 4 data-sampling subprocesses. \emph{Same HGT setting} fixes every implementation settings (see "Implementation Details." in section \ref{sec:exp_setup}), and varies only the data preprocessing pipeline.}
\centering
\begingroup
\def\PanelW{\linewidth}%
\footnotesize
\setlength{\tabcolsep}{3.0pt}
\renewcommand{\arraystretch}{1.12}
\noindent\textbf{(a) Evaluation results (OAG-CS).}\par\smallskip
\resizebox{\PanelW}{!}{%
\begin{tabular}{@{}l|cc|cc|cc@{}}
\hline
\multirow{2}{*}{\makecell[l]{\textbf{Preproc.\ pipeline}}}
& \multicolumn{2}{c|}{\textbf{PF-$L_1$}}
& \multicolumn{2}{c|}{\textbf{PF-$L_2$}}
& \multicolumn{2}{c}{\textbf{PV}} \\
\cline{2-7}
& \textbf{NDCG} & \textbf{MRR} & \textbf{NDCG} & \textbf{MRR} & \textbf{NDCG} & \textbf{MRR} \\
\hline
pyHGT
& $0.718 \pm 0.014$ & $0.823 \pm 0.019$
& $0.403 \pm 0.041$ & $0.439 \pm 0.078$
& $\text{\textbf{0.473}} \pm 0.054$ & $0.288 \pm 0.088$ \\
Our (Same HGT setting)
& $\text{\textbf{0.8510}} \pm 0.013$ & $\text{\textbf{0.8328}}\pm0.003$
& $\text{\textbf{0.4375}} \pm0.009$ & $\text{\textbf{0.5118}} \pm0.002$
& $0.4658\pm0.010$ & $\text{\textbf{0.2932}} \pm0.014$ \\
\hline
\end{tabular}}%
\par\medskip
\noindent\textbf{(b) Resource / latency (OAG-CS, Subproc.\(=4\)).}\par\smallskip
\resizebox{\PanelW}{!}{%
\begin{tabular}{@{}l|r|r|r@{}}
\hline
\textbf{Preproc.\ pipeline} & \textbf{RAM (GB)} & \textbf{Data prep (s/ep.)} & \textbf{Wall (s/ep.)} \\
\hline
pyHGT & $26.86$ & $14.8$ & $219.4$ \\
Our (Same HGT setting) & $4.15$ & $4.9$ & $101.9$ \\
\hline
\end{tabular}}%
\endgroup
\label{tab:preprocess-sensitivity-hgt}
\end{table}

Table~\ref{tab:preprocess-sensitivity-hgt} shows that preprocessing alone
can substantially affect HGT. Our pipeline improves PF-$L_1$/PF-$L_2$
NDCG/MRR while remaining competitive on PV (panel \textbf{(a)}), and it
also cuts peak RAM from \(26.86\) to \(4.15\)\,GB, shortens offline graph
preparation from \(14.8\) to \(4.9\)\,s (\(\sim\!67\%\) reduction
relative to pyHGT), and lowers wall-clock time per training epoch from
\(219.4\) to \(101.9\)\,s when using four data-sampling subprocesses
(\(\sim\!54\%\) reduction; panel \textbf{(b)}). We therefore use our
preprocessing pipeline for all subsequent experiments.

\paragraph{Baseline Models.}
We compare THGFM with ieHGCN~\cite{Yang2020ieHGCN},
SeHGNN~\cite{Yang2023SeHGNN}, HGT~\cite{Hu2020HGT}, and
CTRL~\cite{Li2024CTRL}. ieHGCN and SeHGNN are strong heterogeneous
baselines with efficient semantic aggregation, HGT is the primary
heterogeneous-transformer baseline under the same preprocessing and
training protocol, and CTRL provides a SOTA temporal heterogeneous baseline
that models continuous-time relational influence. Baselines use
hyperparameters and protocols matched to THGFM. Within THGFM, we further
report SSTA-only and RTTA-only branch ablations, and all branch-fusion operators.

\paragraph{Implementation Details.}
\label{parag:impl}

We do not conduct an extensive hyperparameter search, as our goal is to evaluate the effectiveness of the proposed architecture rather than to obtain performance gains through model-specific hyperparameter tuning. To avoid introducing additional tuning confounds, we use a fixed HGT-compatible hyperparameter configuration for all comparable models whenever applicable, covering the hidden dimension, number of layers and attention heads, optimizer, learning rate, dropout, sampling depth and width, and training budget. All models are optimized with AdamW using a learning rate of $10^{-3}$, a dropout rate of $0.2$, and gradient clipping at $0.25$. Training proceeds for $200$ epochs. In each epoch, we uniformly sample without replacement $32$ minibatches of $256$ papers from the training split and traverse these minibatches twice. For each paper, we construct an ego subgraph with a sampling depth of $6$ and a width of $128$. We use a hidden dimension of $256$, $L=3$ layers for all models, and $8$ attention heads for all multi-head attention backbones. To ensure reproducibility, we fix the global random seed to $43$ for Python, NumPy, and PyTorch and enable deterministic CUDA execution, including deterministic cuDNN kernels. All experiments are conducted on a single NVIDIA H100 GPU with $80$~GB of memory.

\subsection{Experimental Results \& Ablation Study}
\label{sec:ablation}

Throughout all ablations, the feature model, training pipeline, and
sampling strategy are held fixed so that differences isolate the graph
operator. Table~\ref{tab:appendix-detailed-ablation-drsf-merged} reports the full results compared with baselines on
OAG-CS, OGBN-MAG, and HTAG under three branch temporal configurations. Reported $\pm$ values are
standard deviations over nine runs, with three runs for each seed in
$\{41,42,43\}$. 

\textbf{Notation.} RTE denotes additive temporal bias on node feature encoding,
and RoTA rotates query/key features before attention scoring. We compare three
\textit{branch temporal configurations} shared across all benchmarks.
\texttt{r001\_s001} enables RTE with RoTA off (RTE-Only); \texttt{r110\_s110}
enables RoTA with RTE off (RoTA-Only); \texttt{r111\_s111} enables both RoTA and RTE.

\begin{table*}[t]
  \caption{Full detailed result grids under three branch temporal configurations on OAG-CS, OGBN-MAG, and HTAG.}
  \centering
  \def\AppAblBundleW{\textwidth}%
  \resizebox{\AppAblBundleW}{!}{%
  \begin{minipage}{\textwidth}
  \begingroup
  \def\PanelW{\linewidth}%
  \scriptsize
  \setlength{\tabcolsep}{2.4pt}
  \renewcommand{\arraystretch}{0.92}
  \noindent\textbf{(a) OAG-CS: baselines and DRSF fusion operators.}\par\smallskip
  \resizebox{\PanelW}{!}{%
  \begin{tabular}{c|c|c|c|c|c|c|c|c|c|c}
  \hline\hline
  \textbf{Models}
  & \makecell[c]{\textbf{Params}\\ \textbf{(M)}}
  & \makecell[c]{\textbf{PF-$L_1$}\\ \textbf{NDCG}}
  & \makecell[c]{\textbf{PF-$L_1$}\\ \textbf{MRR}}
  & \makecell[c]{\textbf{PF-$L_1$}\\ \textbf{Rel. Gain}\\ \textbf{vs.\ HGT (\%)}}
  & \makecell[c]{\textbf{PF-$L_2$}\\ \textbf{NDCG}}
  & \makecell[c]{\textbf{PF-$L_2$}\\ \textbf{MRR}}
  & \makecell[c]{\textbf{PF-$L_2$}\\ \textbf{Rel. Gain}\\ \textbf{vs.\ HGT (\%)}}
  & \makecell[c]{\textbf{PV}\\ \textbf{NDCG}}
  & \makecell[c]{\textbf{PV}\\ \textbf{MRR}}
  & \makecell[c]{\textbf{PV}\\ \textbf{Rel. Gain}\\ \textbf{vs.\ HGT (\%)}} \\
  \hline
  CTRL & 15.40 & $ \text{\underline{0.8522}} \pm0.015$ & $\text{\underline{0.8410}}\pm0.007$ & \underline{$+0.56$} & $0.4041\pm0.003$ & $0.4462\pm0.006$ & $-10.23$ & $0.4850 \pm0.003$ & $\text{\underline{0.3157}} \pm0.009$ & \underline{$+5.90$} \\
  ieHGCN & 9.30 & $0.8266\pm0.005$ & $0.8097\pm0.003$ & $-2.82$ & $0.3748\pm0.003$ & $0.3973\pm0.006$ & $-18.35$ & $\text{\underline{0.4878}} \pm0.005$ & $0.3122\pm0.003$ & $+5.60$ \\
  SeHGNN & 3.80 & $0.8319\pm0.006$ & $0.8221\pm0.012$ & $-1.76$ & $0.3771\pm0.002$ & $0.4035\pm0.005$ & $-17.48$ & $0.4421\pm0.012$ & $0.2658\pm0.002$ & $-7.22$ \\
  HGT & 8.20 & $0.8510\pm0.013$ & $0.8328\pm0.003$ & $0.00$ & $\text{\underline{0.4375}} \pm0.009$ & $\text{\underline{0.5118}} \pm0.002$ & \underline{$0.00$} & $0.4658\pm0.010$ & $0.2932\pm0.014$ & $0.00$ \\
  \hline
  \multicolumn{11}{l}{\textbf{Configuration \texttt{r001\_s001}}} \\
  \hline
  SSTA & 4.45 & $0.8511\pm0.009$ & $0.8359\pm0.011$ & $+0.19$ & $\text{\underline{0.4504}}\pm0.0001$ & $\text{\underline{0.5393}}\pm0.0001$ & \underline{$+4.16$} & $0.4963\pm0.013$ & $0.3185\pm0.009$ & $+7.59$ \\
  RTTA & 7.01 & $0.8528\pm0.007$ & $0.8305\pm0.007$ & $-0.03$ & $0.4189\pm0.007$ & $0.4806\pm0.008$ & $-5.17$ & $0.4864\pm0.001$ & $0.3119\pm0.005$ & $+5.40$ \\
  SF & 8.59 & $0.8542\pm0.001$ & $0.8353\pm0.003$ & $+0.34$ & $0.4284\pm0.0001$ & $0.4979\pm0.0005$ & $-2.40$ & $0.4990\pm0.013$ & $0.3260\pm0.005$ & $+9.16$ \\
  LF & 8.99 & $0.8565\pm0.014$ & $0.8389\pm0.008$ & $+0.69$ & $0.4240\pm0.0007$ & $0.4910\pm0.0013$ & $-3.57$ & $0.4983\pm0.012$ & $0.3237\pm0.008$ & $+8.69$ \\
  NGSF & 9.38 & $\text{\underline{0.8598}}\pm0.006$ & $\text{\underline{0.8435}}\pm0.012$ & \underline{$+1.16$} & $0.4427\pm0.0004$ & $0.5256\pm0.0004$ & $+1.94$ & $\text{\underline{0.4993}}\pm0.015$ & $\text{\underline{0.3279}}\pm0.009$ & \underline{$+9.51$} \\
  NGLF & 9.78 & $0.8580\pm0.014$ & $0.8402\pm0.009$ & $+0.86$ & $0.4411\pm0.0006$ & $0.5217\pm0.0006$ & $+1.38$ & $0.4847\pm0.006$ & $0.3104\pm0.004$ & $+4.96$ \\
  TC-NGSF & 9.43 & $0.8569\pm0.006$ & $0.8385\pm0.011$ & $+0.69$ & $0.4400\pm0.0002$ & $0.5204\pm0.0007$ & $+1.13$ & $0.4963\pm0.014$ & $0.3242\pm0.004$ & $+8.56$ \\
  TC-NGLF & 9.83 & $0.8586\pm0.011$ & $0.8423\pm0.009$ & $+1.02$ & $0.4502\pm0.0003$ & $0.5387\pm0.0005$ & $+4.08$ & $0.4903\pm0.005$ & $0.3157\pm0.013$ & $+6.47$ \\
  \hline
  \multicolumn{11}{l}{\textbf{Configuration \texttt{r110\_s110}}} \\
  \hline
  SSTA & 4.45 & $0.8515\pm0.004$ & $0.8362\pm0.008$ & $+0.23$ & $0.4510\pm0.015$ & $0.5401\pm0.008$ & $+4.31$ & $0.4810\pm0.012$ & $0.3046\pm0.012$ & $+3.58$ \\
  RTTA & 7.01 & $0.8537\pm0.011$ & $0.8382\pm0.007$ & $+0.48$ & $0.4238\pm0.015$ & $0.4881\pm0.010$ & $-3.88$ & $0.4865\pm0.012$ & $0.3120\pm0.004$ & $+5.43$ \\
  SF & 8.59 & $0.8530\pm0.014$ & $0.8296\pm0.014$ & $-0.07$ & $0.4299\pm0.0001$ & $0.5012\pm0.0003$ & $-1.90$ & $\text{\textcolor{red}{\textbf{\underline{0.5102}}}}\pm0.001$ & $\text{\textcolor{red}{\textbf{\underline{0.3378}}}}\pm0.001$ & \textcolor{red}{\textbf{\underline{$+12.37$}}} \\
  LF & 8.99 & $0.8565\pm0.006$ & $0.8389\pm0.014$ & $+0.69$ & $0.4072\pm0.0003$ & $0.4548\pm0.0015$ & $-9.03$ & $0.4983\pm0.010$ & $0.3237\pm0.012$ & $+8.69$ \\
  NGSF & 9.38 & $\text{\textcolor{red}{\textbf{\underline{0.8603}}}}\pm0.008$ & $0.8426\pm0.009$ & $+1.13$ & $\text{\underline{0.4512}}\pm0.0004$ & $\text{\underline{0.5410}}\pm0.0015$ & \underline{$+4.42$} & $0.4954\pm0.005$ & $0.3225\pm0.003$ & $+8.17$ \\
  NGLF & 9.78 & $0.8577\pm0.012$ & $0.8427\pm0.009$ & $+0.99$ & $0.4407\pm0.0002$ & $0.5198\pm0.0013$ & $+1.15$ & $0.4858\pm0.003$ & $0.3116\pm0.012$ & $+5.28$ \\
  TC-NGSF & 9.43 & $0.8597\pm0.014$ & $\text{\textcolor{red}{\textbf{\underline{0.8439}}}}\pm0.004$ & \textcolor{red}{\textbf{\underline{$+1.18$}}} & $0.4421\pm0.005$ & $0.5245\pm0.003$ & $+1.77$ & $0.5008\pm0.007$ & $0.3285\pm0.004$ & $+9.78$ \\
  TC-NGLF & 9.83 & $0.8570\pm0.005$ & $0.8398\pm0.015$ & $+0.77$ & $0.4305\pm0.007$ & $0.5043\pm0.006$ & $-1.53$ & $0.4904\pm0.004$ & $0.3262\pm0.009$ & $+8.27$ \\
  \hline
  \multicolumn{11}{l}{\textbf{Configuration \texttt{r111\_s111}}} \\
  \hline
  SSTA & 4.45 & $0.8566\pm0.002$ & $0.8407\pm0.004$ & $+0.80$ & $0.4372\pm0.0001$ & $0.5160\pm0.0005$ & $+0.38$ & $0.4983\pm0.007$ & $0.3155\pm0.004$ & $+7.29$ \\
  RTTA & 7.01 & $0.8531\pm0.003$ & $0.8386\pm0.004$ & $+0.47$ & $0.4254\pm0.007$ & $0.4903\pm0.002$ & $-3.48$ & $0.4976\pm0.013$ & $0.3235\pm0.015$ & $+8.58$ \\
  SF & 8.59 & $0.8559\pm0.004$ & $0.8403\pm0.009$ & $+0.74$ & $0.4274\pm0.007$ & $0.4964\pm0.008$ & $-2.66$ & $\text{\underline{0.5034}}\pm0.004$ & $\text{\underline{0.3298}}\pm0.003$ & \underline{$+10.28$} \\
  LF & 8.99 & $0.8559\pm0.011$ & $0.8414\pm0.012$ & $+0.80$ & $0.4233\pm0.0005$ & $0.4886\pm0.0010$ & $-3.89$ & $0.4998\pm0.002$ & $0.3259\pm0.003$ & $+9.23$ \\
  NGSF & 9.38 & $0.8587\pm0.014$ & $0.8431\pm0.003$ & $+1.07$ & $0.4428\pm0.0001$ & $0.5267\pm0.0003$ & $+2.06$ & $0.4858\pm0.010$ & $0.3139\pm0.012$ & $+5.68$ \\
  NGLF & 9.78 & $0.8526\pm0.009$ & $0.8302\pm0.001$ & $-0.06$ & $\text{\textcolor{red}{\textbf{\underline{0.4528}}}}\pm0.0003$ & $\text{\textcolor{red}{\textbf{\underline{0.5437}}}}\pm0.0006$ & \textcolor{red}{\textbf{\underline{$+4.87$}}} & $0.4879\pm0.005$ & $0.3139\pm0.004$ & $+5.90$ \\
  TC-NGSF & 9.43 & $\text{\underline{0.8600}}\pm0.001$ & $\text{\underline{0.8437}}\pm0.001$ & \textcolor{red}{\textbf{\underline{$+1.18$}}} & $0.4394\pm0.014$ & $0.5177\pm0.013$ & $+0.79$ & $0.4967\pm0.004$ & $0.3252\pm0.002$ & $+8.77$ \\
  TC-NGLF & 9.83 & $0.8569\pm0.009$ & $0.8409\pm0.010$ & $+0.83$ & $0.4397\pm0.002$ & $0.5190\pm0.008$ & $+0.95$ & $0.4936\pm0.003$ & $0.3214\pm0.010$ & $+7.79$ \\
  \hline\hline
  \end{tabular}%
  }%
  \par\medskip
  \renewcommand{\arraystretch}{0.92}
  \noindent\textbf{(b) OGBN-MAG and HTAG: baselines and DRSF fusion operators.}\par\smallskip
  \resizebox{\PanelW}{!}{%
  \begin{tabular}{c|c|c|c|c|c|c|c|c|c|c}
  \hline\hline
  \textbf{Models}
  & \makecell[c]{\textbf{Params}\\ \textbf{(M)}}
  & \makecell[c]{\textbf{OGBN-MAG(PV)}\\ \textbf{Test Acc.}}
  & \makecell[c]{\textbf{OGBN-MAG(PV)}\\ \textbf{Val. Acc.}}
  & \makecell[c]{\textbf{OGBN-MAG}\\ \textbf{Rel. Gain}\\ \textbf{vs.\ HGT (\%)} }
  & \makecell[c]{\textbf{HTAG-ArXiv}\\ \textbf{Test Acc.}}
  & \makecell[c]{\textbf{HTAG-ArXiv}\\ \textbf{Val. Acc.}}
  & \makecell[c]{\textbf{HTAG-ArXiv}\\ \textbf{Rel. Gain}\\ \textbf{vs.\ HGT (\%)} }
  & \makecell[c]{\textbf{HTAG-DBLP}\\ \textbf{Test Acc.}}
  & \makecell[c]{\textbf{HTAG-DBLP}\\ \textbf{Val. Acc.}}
  & \makecell[c]{\textbf{HTAG-DBLP}\\ \textbf{Rel. Gain}\\ \textbf{vs.\ HGT (\%)} } \\
  \hline
CTRL & 15.40 & $\text{\underline{0.4017}} \pm 0.007$ & $0.4019 \pm 0.008$ & $-0.82$ & $\text{\underline{0.8443}} \pm 0.010$ & $\text{\underline{0.8415}} \pm 0.003$ & \underline{$+2.08$} & $\text{\underline{0.7540}} \pm 0.014$ & $0.7634 \pm 0.005$ & \underline{$+0.53$} \\
 ieHGCN & 9.30 & $0.3703 \pm 0.008$ & $0.3879 \pm 0.014$ & $-6.51$ & $0.8232 \pm 0.012$ & $0.8115 \pm 0.013$ & $-1.02$ & $0.7305 \pm 0.009$ & $0.7651 \pm 0.014$ & $-0.92$ \\
  SeHGNN & 3.80 & $0.3694 \pm 0.009$ & $0.3684 \pm 0.012$ & $-8.94$ & $0.8281 \pm 0.013$ & $0.8289 \pm 0.012$ & $+0.34$ & $0.7383 \pm 0.011$ & $\text{\underline{0.7656}} \pm 0.020$ & $-0.37$ \\
  HGT & 8.20 & $0.3906 \pm 0.012$ & $\text{\underline{0.4208}} \pm 0.020$ & \underline{$+0.00$} & $0.8352 \pm 0.014$ & $0.8164 \pm 0.005$ & $+0.00$ & $0.7508 \pm 0.029$ & $0.7586 \pm 0.015$ & $+0.00$ \\
  \hline
  \multicolumn{11}{l}{\textbf{Configuration \texttt{r001\_s001}}} \\
  \hline
 SSTA & 4.45 & $0.4006 \pm 0.019$ & $0.4175 \pm 0.014$ & $+0.89$ & $0.8277 \pm 0.011$ & $0.8246 \pm 0.014$ & $+0.05$ & $0.7645 \pm 0.013$ & $0.7728 \pm 0.009$ & $+1.85$ \\
  RTTA & 7.01 & $0.4001 \pm 0.013$ & $0.4176 \pm 0.026$ & $+0.84$ & $0.8539 \pm 0.009$ & $0.8430 \pm 0.014$ & $+2.75$ & $0.7848 \pm 0.012$ & $0.7754 \pm 0.012$ & $+3.37$ \\
  SF & 8.59 & $0.3992 \pm 0.008$ & $0.4238 \pm 0.005$ & $+1.46$ & $\text{\underline{0.8570}} \pm 0.011$ & $0.8430 \pm 0.017$ & \underline{$+2.93$} & $0.7699 \pm 0.007$ & $0.7656 \pm 0.008$ & $+1.73$ \\
  LF & 8.99 & $0.3952 \pm 0.008$ & $0.4105 \pm 0.010$ & $-0.64$ & $0.8418 \pm 0.014$ & $0.8434 \pm 0.012$ & $+2.05$ & $0.7617 \pm 0.013$ & $0.7715 \pm 0.013$ & $+1.58$ \\
  NGSF & 9.38 & $\text{\underline{0.4036}} \pm 0.022$ & $\text{\underline{0.4249}} \pm 0.011$ & \underline{$+2.15$} & $0.8449 \pm 0.012$ & $0.8426 \pm 0.014$ & $+2.19$ & $0.7797 \pm 0.012$ & $0.7695 \pm 0.007$ & $+2.64$ \\
  NGLF & 9.78 & $0.3865 \pm 0.011$ & $0.4042 \pm 0.010$ & $-2.50$ & $0.8359 \pm 0.014$ & $0.8406 \pm 0.017$ & $+1.52$ & $0.7570 \pm 0.013$ & $0.7637 \pm 0.013$ & $+0.75$ \\
  TC-NGSF & 9.43 & $0.3982 \pm 0.013$ & $0.4232 \pm 0.009$ & $+1.26$ & $0.8316 \pm 0.011$ & $0.8379 \pm 0.007$ & $+1.10$ & $\text{\underline{0.7863}} \pm 0.015$ & $\text{\underline{0.7820}} \pm 0.007$ & \underline{$+3.91$} \\
  TC-NGLF & 9.83 & $0.3808 \pm 0.015$ & $0.4047 \pm 0.010$ & $-3.17$ & $0.8441 \pm 0.009$ & $\text{\underline{0.8438}} \pm 0.010$ & $+2.21$ & $0.7660 \pm 0.010$ & $0.7652 \pm 0.012$ & $+1.45$ \\
  \hline
  \multicolumn{11}{l}{\textbf{Configuration \texttt{r110\_s110}}} \\
  \hline
 SSTA & 4.45 & $0.3878 \pm 0.015$ & $0.4049 \pm 0.018$ & $-2.25$ & $0.7766 \pm 0.012$ & $0.7773 \pm 0.015$ & $-5.90$ & $0.7398 \pm 0.012$ & $0.7527 \pm 0.015$ & $-1.12$ \\
  RTTA & 7.01 & $0.4002 \pm 0.013$ & $0.4188 \pm 0.024$ & $+0.99$ & $0.8301 \pm 0.017$ & $0.8324 \pm 0.015$ & $+0.67$ & $0.7840 \pm 0.009$ & $0.7758 \pm 0.012$ & $+3.34$ \\
  SF & 8.59 & $0.3998 \pm 0.005$ & $0.4248 \pm 0.015$ & $+1.65$ & $\text{\underline{0.8570}} \pm 0.009$ & $0.8441 \pm 0.017$ & \underline{$+3.00$} & $0.7848 \pm 0.017$ & $0.7707 \pm 0.007$ & $+3.06$ \\
  LF & 8.99 & $0.3847 \pm 0.014$ & $0.4078 \pm 0.008$ & $-2.30$ & $0.8418 \pm 0.013$ & $0.8406 \pm 0.017$ & $+1.88$ & $0.7672 \pm 0.016$ & $0.7711 \pm 0.007$ & $+1.92$ \\
  NGSF & 9.38 & $\text{\underline{0.4127}} \pm 0.015$ & $\text{\textcolor{red}{\textbf{\underline{0.4327}}}} \pm 0.015$ & \textcolor{red}{\underline{$+4.24$}} & $0.8430 \pm 0.015$ & $0.8258 \pm 0.014$ & $+1.04$ & $0.7871 \pm 0.008$ & $0.7754 \pm 0.007$ & $+3.52$ \\
  NGLF & 9.78 & $0.3804 \pm 0.015$ & $0.4056 \pm 0.016$ & $-3.11$ & $0.8469 \pm 0.022$ & $0.8352 \pm 0.008$ & $+1.85$ & $0.7807 \pm 0.001$ & $\text{\textcolor{red}{\textbf{\underline{0.7984}}}} \pm 0.035$ & \textcolor{red}{\textbf{\underline{$+4.61$}}} \\
  TC-NGSF & 9.43 & $0.4065 \pm 0.012$ & $0.4257 \pm 0.016$ & $+2.62$ & $0.8406 \pm 0.013$ & $0.8430 \pm 0.011$ & $+1.95$ & $\text{\underline{0.7875}} \pm 0.013$ & $0.7754 \pm 0.007$ & $+3.55$ \\
  TC-NGLF & 9.83 & $0.3838 \pm 0.020$ & $0.4079 \pm 0.022$ & $-2.40$ & $0.8398 \pm 0.007$ & $\text{\underline{0.8473}} \pm 0.015$ & $+2.17$ & $0.7852 \pm 0.013$ & $0.7789 \pm 0.006$ & $+3.63$ \\
  \hline
  \multicolumn{11}{l}{\textbf{Configuration \texttt{r111\_s111}}} \\
  \hline
 SSTA & 4.45 & $0.4017 \pm 0.018$ & $0.4166 \pm 0.017$ & $+0.92$ & $0.7660 \pm 0.013$ & $0.7723 \pm 0.016$ & $-6.84$ & $0.7473 \pm 0.008$ & $0.7621 \pm 0.009$ & $+0.00$ \\
  RTTA & 7.01 & $0.3999 \pm 0.013$ & $0.4163 \pm 0.013$ & $+0.66$ & $0.8391 \pm 0.009$ & $0.8367 \pm 0.015$ & $+1.48$ & $0.7766 \pm 0.012$ & $0.7766 \pm 0.007$ & $+2.90$ \\
  SF & 8.59 & $0.4012 \pm 0.018$ & $0.4246 \pm 0.022$ & $+1.81$ & $0.8544 \pm 0.002$ & $\text{\textcolor{red}{\textbf{\underline{0.8586}}}} \pm 0.026$ & \textcolor{red}{\textbf{\underline{$+3.73$}}} & $0.7742 \pm 0.013$ & $0.7594 \pm 0.013$ & $+1.61$ \\
  LF & 8.99 & $0.3958 \pm 0.014$ & $0.4156 \pm 0.016$ & $+0.05$ & $0.8297 \pm 0.013$ & $0.8293 \pm 0.013$ & $+0.46$ & $0.7730 \pm 0.013$ & $0.7578 \pm 0.010$ & $+1.43$ \\
  NGSF & 9.38 & $0.4046 \pm 0.017$ & $\text{\underline{0.4265}} \pm 0.018$ & $+2.47$ & $0.8387 \pm 0.010$ & $0.8344 \pm 0.013$ & $+1.31$ & $\text{\textcolor{red}{\textbf{\underline{0.7895}}}} \pm 0.010$ & $0.7727 \pm 0.013$ & $+3.51$ \\
  NGLF & 9.78 & $0.3785 \pm 0.012$ & $0.3955 \pm 0.009$ & $-4.56$ & $0.8270 \pm 0.006$ & $0.8313 \pm 0.011$ & $+0.42$ & $0.7836 \pm 0.008$ & $0.7777 \pm 0.013$ & $+3.44$ \\
  TC-NGSF & 9.43 & $\text{\textcolor{red}{\textbf{\underline{0.4150}}}} \pm 0.005$ & $0.4253 \pm 0.016$ & \textbf{\underline{$+3.66$}} & $0.8531 \pm 0.007$ & $0.8414 \pm 0.015$ & $+2.60$ & $0.7879 \pm 0.011$ & $0.7809 \pm 0.008$ & \underline{$+3.94$} \\
  TC-NGLF & 9.83 & $0.3715 \pm 0.017$ & $0.3981 \pm 0.016$ & $-5.14$ & $\text{\textcolor{red}{\textbf{\underline{0.8598}}}} \pm 0.019$ & $0.8422 \pm 0.009$ & $+3.05$ & $0.7754 \pm 0.012$ & $\text{\underline{0.7816}} \pm 0.015$ & $+3.15$ \\
  \hline\hline
  \end{tabular}%
  }%
  \endgroup
  \end{minipage}%
  }%
  \label{tab:appendix-detailed-ablation-drsf-merged}
  \end{table*}

To summarize models on cross-dataset behavior, we aggregate OAG-CS (PF-$L_1$,
PF-$L_2$, PV), OGBN-MAG (PV), HTAG-ArXiv, and HTAG-DBLP into a
\emph{six-task mean gain} against HGT: for each task, we take the
\emph{Relative Gain (\%) vs.\ HGT} reported in
Table~\ref{tab:appendix-detailed-ablation-drsf-merged} and average the six
tasks with equal weight. The same table also lists all result cells under
each branch temporal configuration; Table~\ref{tab:cross_task_fusion_by_family} gives a compact
structural summary.

\begin{table}[t]
\caption{Cross-dataset fusion summarized by six-task mean gain vs.\ HGT.}
\centering
\begingroup
\def\PanelW{\linewidth}%
\scriptsize
\renewcommand{\arraystretch}{1.08}
\setlength{\tabcolsep}{3pt}
\noindent\textbf{(a) Top three fusion operators for each fixed branch temporal configuration.}\par\smallskip
\resizebox{\PanelW}{!}{%
\begin{tabular}{@{}l c r r c r r c r r c@{}}
\toprule
\textbf{Config.}
& \multicolumn{3}{c}{\textbf{Rank 1}}
& \multicolumn{3}{c}{\textbf{Rank 2}}
& \multicolumn{3}{c}{\textbf{Rank 3}}
& \makecell{\textbf{Overall Mean Gain}\\\textbf{vs.\ HGT (\%)}} \\
\cmidrule(lr){2-4} \cmidrule(lr){5-7} \cmidrule(lr){8-10} \cmidrule(lr){11-11}
& \textbf{Fusion Op.} & \makecell{\textbf{P.}\\\textbf{(M)}} & \makecell{\textbf{Mean}\\\textbf{(\%)}} & \textbf{Fusion Op.} & \makecell{\textbf{P.}\\\textbf{(M)}} & \makecell{\textbf{Mean}\\\textbf{(\%)}} & \textbf{Fusion Op.} & \makecell{\textbf{P.}\\\textbf{(M)}} & \makecell{\textbf{Mean}\\\textbf{(\%)}} & \makecell{} \\
\midrule
\texttt{r001\_s001} & NGSF & 9.38 & $+3.27$ & TC-NGSF & 9.43 & $+2.78$ & SSTA & 4.45 & $+2.46$ & $+2.83$ \\
\textcolor{red}{\texttt{r110\_s110}} & NGSF & 9.38 & $+3.75$ & TC-NGSF & 9.43 & $+3.48$ & SF & 8.59 & $+3.02$ & $\textcolor{red}{\textbf{+3.42}}$ \\
\texttt{r111\_s111} & TC-NGSF & 9.43 & $+3.49$ & NGSF & 9.38 & $+2.68$ & SF & 8.59 & $+2.59$ & $+2.92$ \\
\bottomrule
\end{tabular}}%
\par\medskip
\renewcommand{\arraystretch}{1.05}
\noindent\textbf{(b) Per-task gains for fusion pools and for each task's best single (operator, configuration).}\par\smallskip
\resizebox{\PanelW}{!}{%
\begin{tabular}{@{}l c c c c c c c@{}}
\toprule
\textbf{(Pool of) fusion operators}
& \makecell{\textbf{OAG-CS}\\\textbf{PF-$L_1$}}
& \makecell{\textbf{OAG-CS}\\\textbf{PF-$L_2$}}
& \makecell{\textbf{OAG-CS}\\\textbf{PV}}
& \makecell{\textbf{OGBN-}\\\textbf{MAG}}
& \makecell{\textbf{HTAG-}\\\textbf{ArXiv}}
& \makecell{\textbf{HTAG-}\\\textbf{DBLP}}
& \makecell{\textbf{Mean Gain}\\\textbf{vs.\ HGT (\%)}} \\
\midrule
SF & $+0.34$ & $-2.32$ & $+10.60$ & $+1.64$ & $+3.22$ & $+2.13$ & $+2.60$ \\
LF & $+0.73$ & $-5.50$ & $+8.87$ & $-0.96$ & $+1.46$ & $+1.64$ & $+1.04$ \\
NGSF & $+1.12$ & $+2.81$ & $+7.79$ & $+2.95$ & $+1.51$ & $+3.22$ & $+3.23$ \\
NGLF & $+0.60$ & $+2.47$ & $+5.38$ & $-3.39$ & $+1.26$ & $+2.93$ & $+1.54$ \\
\textcolor{red}{TC-NGSF} & $+1.02$ & $+1.23$ & $+9.04$ & $+2.51$ & $+1.88$ & $+3.80$ & $\textcolor{red}{\textbf{+3.25}}$ \\
TC-NGLF & $+0.87$ & $+1.17$ & $+7.51$ & $-3.57$ & $+2.48$ & $+2.74$ & $+1.87$ \\
\hline
Plain (SF, LF)
& $+0.53$ & $-3.91$ & $+9.74$ & $+0.34$ & $+2.34$ & $+1.89$ & $+1.82$ \\
NG$^\ast$ (NGSF, NGLF)
& $+0.86$ & $+2.64$ & $+6.58$ & $-0.22$ & $+1.39$ & $+3.08$ & $+2.39$ \\
\textcolor{red}{TC-NG$^\ast$ (TC-NGSF, TC-NGLF)}
& $+0.95$ & $+1.20$ & $+8.27$ & $-0.53$ & $+2.18$ & $+3.27$ & $\textcolor{red}{\textbf{+2.56}}$ \\
\hline
\textcolor{red}{Sum-style (SF, NGSF, TC-NGSF)}
& $+0.82$ & $+0.57$ & $+9.14$ & $+2.37$ & $+2.21$ & $+3.05$ & $\textcolor{red}{\textbf{+3.03}}$ \\
Linear-style (LF, NGLF, TC-NGLF)
& $+0.73$ & $-0.62$ & $+7.25$ & $-2.64$ & $+1.73$ & $+2.44$ & $+1.48$ \\
\hline
\makecell[l]{Best single\\(operator, configuration)}
& \makecell[c]{TC-NGSF@ \\ \texttt{r111\_s111}\\$+1.18$}
& \makecell[c]{NGLF@ \\ \texttt{r111\_s111}\\$+4.87$}
& \makecell[c]{SF@ \\ \texttt{r110\_s110}\\$+12.37$}
& \makecell[c]{NGSF@ \\ \texttt{r110\_s110}\\$+4.24$}
& \makecell[c]{SF@ \\ \texttt{r111\_s111}\\$+3.73$}
& \makecell[c]{NGLF@ \\ \texttt{r110\_s110}\\$+4.61$}
& $+5.17$ \\
\bottomrule
\end{tabular}}%
\endgroup
\label{tab:cross_task_fusion_by_family}
\end{table}

Table~\ref{tab:cross_task_fusion_by_family}\textbf{(a)} fixes each branch
temporal configuration and ranks the top three fusion operators by six-task mean gain.
NGSF ranks first under RTE-only and under
RoTA-only, while TC-NGSF ranks first when both
are enabled (\texttt{r111\_s111}). Averaging the top-three entries in each row,
\textbf{RoTA-only} attains the highest mean ($+3.42\%$), clearly
above RTE-only ($+2.83\%$), so RoTA contributes more than
RTE in this aggregate view. Enabling RoTA and RTE together lowers the mean to
$+2.92\%$, i.e., a drop from the RoTA-only peak, but this joint setting still
outperforms RTE-only ($+2.92\%$ vs.\ $+2.83\%$).

Table~\ref{tab:cross_task_fusion_by_family}\textbf{(b)} average fusion operators inside
each pool, then averages over the three configurations. 
Under this aggregation, six-task pool-mean gains increase monotonically from
Plain ($+1.82\%$) to NG$^\ast$ ($+2.39\%$) to
\textbf{TC-NG$^\ast$} ($+2.56\%$), shows the \textbf{TC-NG$^\ast$} pool improves
reliably over NG$^\ast$ and over ungated sum/linear fusion.
Separately, \textbf{sum-style} pooling (SF, NGSF, TC-NGSF) at
$+3.03\%$ beats linear-style pooling (LF, NGLF, TC-NGLF) at
$+1.48\%$, shows sum-style fusion outperform linear-style fusion on the
cross-task average in this grid.

\subsection{Discussion}
\label{sec:discussion}

\begin{table}[t]
  \caption{Per-layer compute complexity.
  \(N\) and \(M\) are the number of nodes and edges in a
  sampled batch subgraph, \(d\) is the hidden dimension, \(d_{\tau}\)
  the type-embedding dimension, and
  \(|\mathcal{T}|\), \(|\mathcal{R}|\) are the numbers of node types and
  relation types.}
  \centering
  \def\PanelW{\linewidth}%
  \begingroup
  \footnotesize
  \setlength{\tabcolsep}{3.2pt}
  \renewcommand{\arraystretch}{1.08}
  \def\CostSSTA{\mathcal{O}(N d^2 + M d)}%
  \def\CostRTTA{\mathcal{O}\!\bigl(N d^2 + M d + M(|\mathcal{T}|+|\mathcal{R}|)\bigr)}%
  \resizebox{0.6\PanelW}{!}{%
  \begin{tabular}{@{}l|l@{}}
  \hline
  \textbf{Module / operator} & \textbf{Compute complexity (per layer)} \\
  \hline
  (Baseline) HGT
  & \(\mathcal{O}(N d^2 + M d + M|\mathcal{T}|^2|\mathcal{R}|)\) \\
  \hline
  \multicolumn{2}{@{}l@{}}{\emph{Single-branch}} \\
  \hline
  SSTA
  & \(\CostSSTA\) \\
  RTTA
  & \(\CostRTTA\) \\
  \hline
  \multicolumn{2}{@{}l@{}}{\emph{THGFM}} \\
  \hline
  SF & \(SSTA + RTTA + \mathcal{O}(N d)\) \\
  LF & \(SSTA + RTTA + \mathcal{O}(N d^2)\) \\
  NGLF & \(SSTA + RTTA + \mathcal{O}(N d^2)\) \\
  NGSF & \(SSTA + RTTA + \mathcal{O}(N d^2)\) \\
  TC-NGLF & \(SSTA + RTTA + \mathcal{O}(N d^2 + N d d_{\tau})\) \\
  TC-NGSF & \(SSTA + RTTA + \mathcal{O}(N d^2 + N d d_{\tau})\) \\
  \hline
  \end{tabular}}%
  \endgroup
  \label{tab:discussion-complexity}
\end{table}

Table~\ref{tab:preprocess-sensitivity-hgt} is intended to isolate the contribution of preprocessing. It keeps the HGT architecture and training configuration fixed while varying only the preprocessing pipeline, thereby directly quantifying the gains attributable to preprocessing. In contrast, the HGT results reported in Tables~\ref{tab:appendix-detailed-ablation-drsf-merged} and~\ref{tab:cross_task_fusion_by_family} already incorporate our preprocessing pipeline. Consequently, the improvements of THGFM in these tables are measured against a stronger HGT baseline under identical preprocessing, feature construction, sampling, and training protocols, thus isolating the gains from the proposed model components and operators. The complete experimental grids in Table~\ref{tab:appendix-detailed-ablation-drsf-merged} further substantiate the contribution of RTTA. Moreover, comparing the per-layer computational hierarchy in Table~\ref{tab:discussion-complexity} with the six-task mean gains in Table~\ref{tab:cross_task_fusion_by_family}\textbf{(b)} reveals two distinct performance--efficiency trade-off directions. \emph{(i) Gating direction:}
moving Plain \(\to\) NG\(^\ast\) \(\to\) TC-NG\(^\ast\) adds only a
\(d_{\tau}\)-dimensional conditioning (\(+0.05\)M parameters for
TC-NG\(^\ast\) over NG\(^\ast\)) but lifts mean gain monotonically from
\(+1.82\%\) to \(+2.39\%\) to \(+2.56\%\); type-conditioning is thus
effectively free while reliably positive. \emph{(ii) Aggregation
direction:} plain SF adds only \(\mathcal{O}(N d)\), while LF adds
\(\mathcal{O}(N d^2)\); after NG/TC gating, sum-style and
linear-style variants have the same leading
\(\mathcal{O}(N d^2)\) fusion order. Yet sum-style yields
\(+3.03\%\) vs.\ \(+1.48\%\) for linear-style, so allocating the
fusion budget to gated summation is more effective than linear mixing
in this regime. The intersection of the two
directions, \textbf{TC-NGSF}, attains the best single-operator mean
gain (\(+3.25\%\)) at only \(9.43\)M parameters, which is cheaper than
both TC-NGLF (\(9.83\)M) and NGLF (\(9.78\)M). Finally, because the
temporal switches have identical parameter and computational complexity,
this comparison isolates the temporal operator itself:
\textbf{RoTA-only} (\texttt{r110\_s110}) gives the highest aggregate
gain (\(+3.42\%\)), clearly above \textbf{RTE-only}
(\texttt{r001\_s001}, \(+2.83\%\)). Thus,
\textbf{TC-NGSF under RoTA-only} emerges as the
most suitable THGFM configuration: it lies at the knee of the
accuracy--cost curve, and outperforms both plain and linear-style fusion on the cross-task average.

\section{Conclusion}
\label{sec:conclusion}

We proposed THGFM, a dual-branch temporal heterogeneous graph
fusion model that pairs a shared-space branch (SSTA) with a
relation-partitioned branch (RTTA), fuses them through a
type-conditioned non-competitive gated sum mechanism (TC-NGSF),
and injects relative time into attention via rotary encoding
(RoTA). THGFM consistently outperforms HGT on OAG-CS, OGBN-MAG,
HTAG-ArXiv, and HTAG-DBLP, reaching up to \(+12.37\%\) on
OAG-CS PV and a \(+3.25\%\) six-task mean gain. Three directions remain open: (i) developing globally consistent
node-time assignment schemes compatible with mini-batch sampling,
to address the induced-time inconsistency discussed in
Section~\ref{sec:preliminaries}; (ii) scaling THGFM into a
temporal heterogeneous graph foundation
model~\cite{BechlerSpeicher2026GraphBFF} pretrained across
more other graphs; and (iii) applying THGFM to Relation Deep Learning.

%
%
%
\bibliographystyle{splncs04}
\bibliography{references}

\end{document}